\documentclass[10pt, conference, compsocconf]{IEEEtran}
\ifCLASSINFOpdf
\else
\fi

\usepackage{subcaption}
\usepackage{url}
\usepackage[export]{adjustbox}
\usepackage{booktabs} 
\usepackage{tikz}

\newcommand\copyrighttext{%
	\footnotesize Copyright 2019 IEEE. Published as a conference paper at the 2019 IEEE Conference 
	on Multimedia Information Processing and Retrieval (MIPR2019) DOI: 
	\url{https://doi.org/10.1109/MIPR.2019.00085}}
\newcommand\copyrightnotice{%
	\begin{tikzpicture}[remember picture,overlay]
	\node[anchor=south,yshift=10pt] at (current page.south) 
	{\fbox{\parbox{\dimexpr\textwidth-\fboxsep-\fboxrule\relax}{\copyrighttext}}};
	\end{tikzpicture}%
}

\begin{document}
%
\title{Image Captioning with Clause-Focused Metrics in a Multi-Modal Setting for Marketing}


\author{\IEEEauthorblockN{Philipp Harzig, Dan Zecha, Rainer Lienhart}
\IEEEauthorblockA{Multimedia Computing and Computer Vision Lab\\
University of Augsburg\\
86159 Augsburg, Germany\\
\{philipp.harzig, dan.zecha, \\ rainer.lienhart\}@informatik.uni-augsburg.de}
\and
\IEEEauthorblockN{Carolin Kaiser, Ren\'{e} Schallner}
\IEEEauthorblockA{GfK Verein\\
90419 Nuremberg, Germany\\
\{carolin.kaiser, rene.schallner \} \\@gfk-verein.org}
}


%


\maketitle

\begin{abstract}
Automatically generating descriptive captions for images is a well-researched area in computer vision.
However, existing evaluation approaches focus on measuring the similarity between two sentences disregarding fine-grained semantics of the captions.
In our setting of images depicting persons interacting with branded products, the subject, predicate, object and the name of the branded product are important evaluation criteria of the generated captions. Generating image captions with these constraints is a new challenge, which we tackle in this work. By simultaneously predicting integer-valued ratings that describe attributes of the human-product interaction, we optimize a deep neural network architecture in a multi-task learning setting, which considerably improves the caption quality. Furthermore, we introduce a novel metric that allows us to assess whether the generated captions meet our requirements (i.e., subject, predicate, object, and product name) and describe a series of experiments on caption quality and how to address annotator disagreements for the image ratings with an approach called soft targets. 
We also show that our novel clause-focused metrics are also applicable to other image captioning datasets, such as the popular MSCOCO dataset.
\end{abstract}


%
\IEEEpeerreviewmaketitle

\copyrightnotice
\section{Introduction}
\begin{figure}[!htb]
	\centering
	\begin{adjustbox}{minipage=\linewidth}
		\begin{subfigure}{.495\linewidth}
			\includegraphics[width=\linewidth]{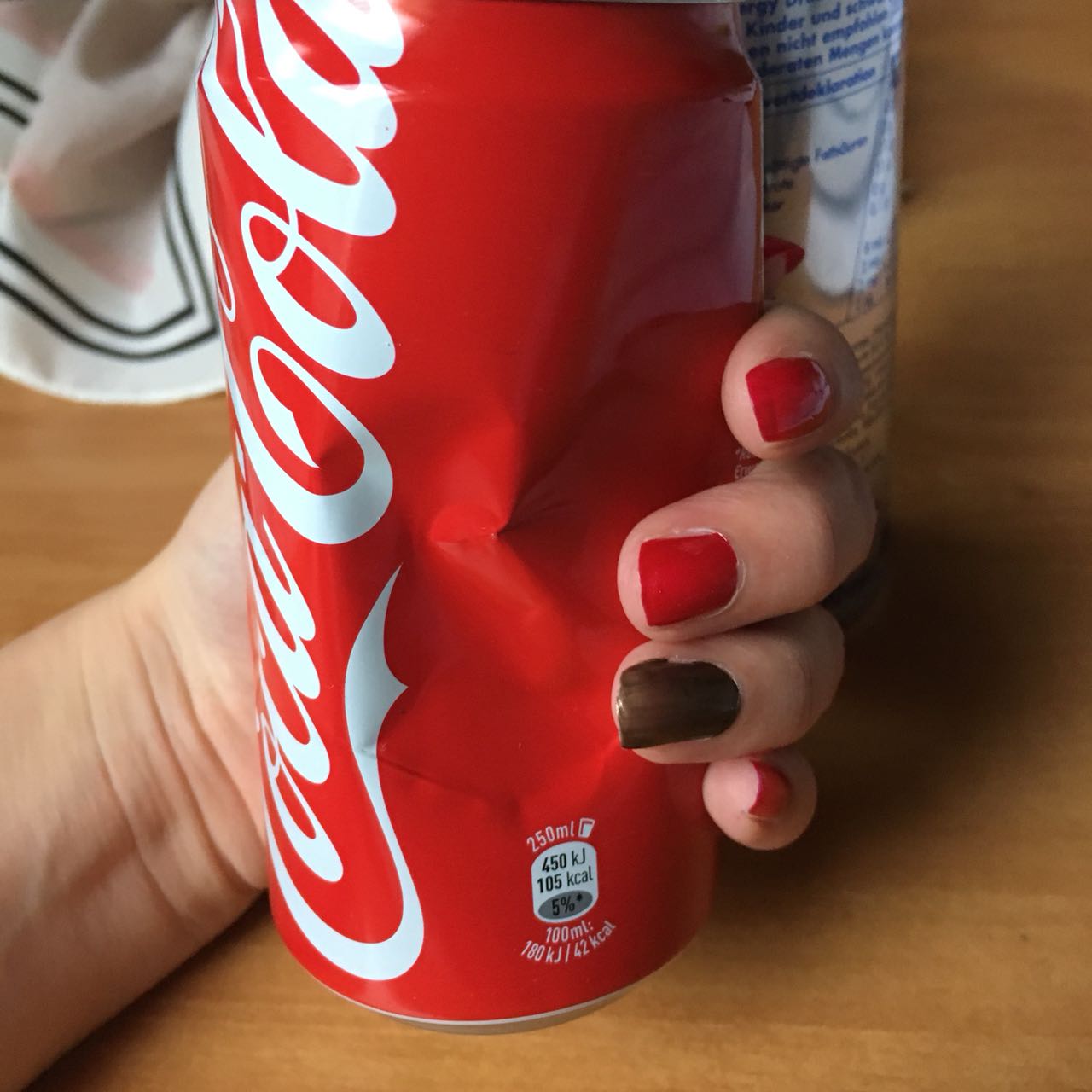}
		\end{subfigure}
		\begin{subfigure}{.495\linewidth}
			\includegraphics[width=\linewidth]{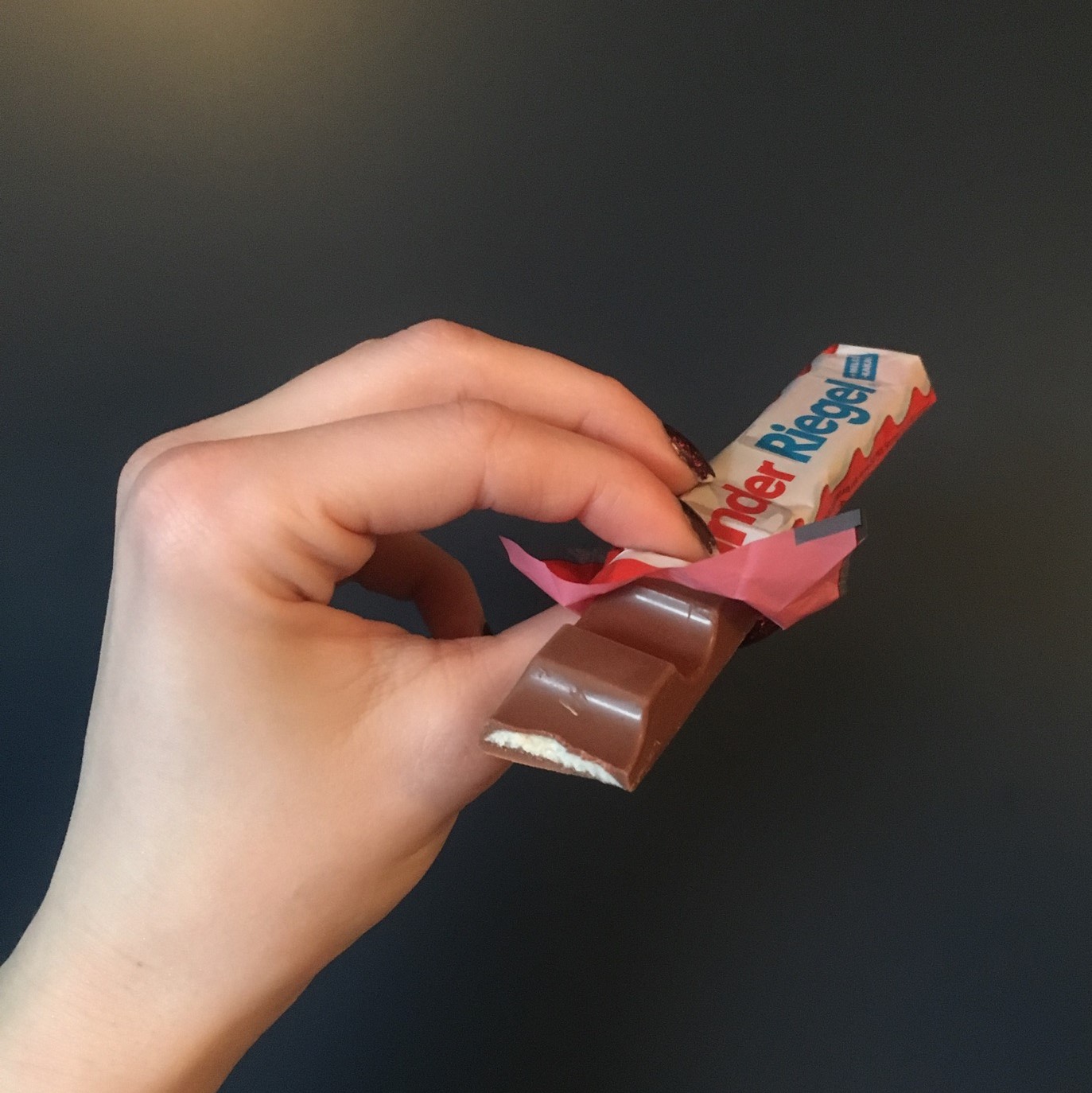}
		\end{subfigure}
		\begin{subfigure}{.495\linewidth}
			\includegraphics[width=\linewidth]{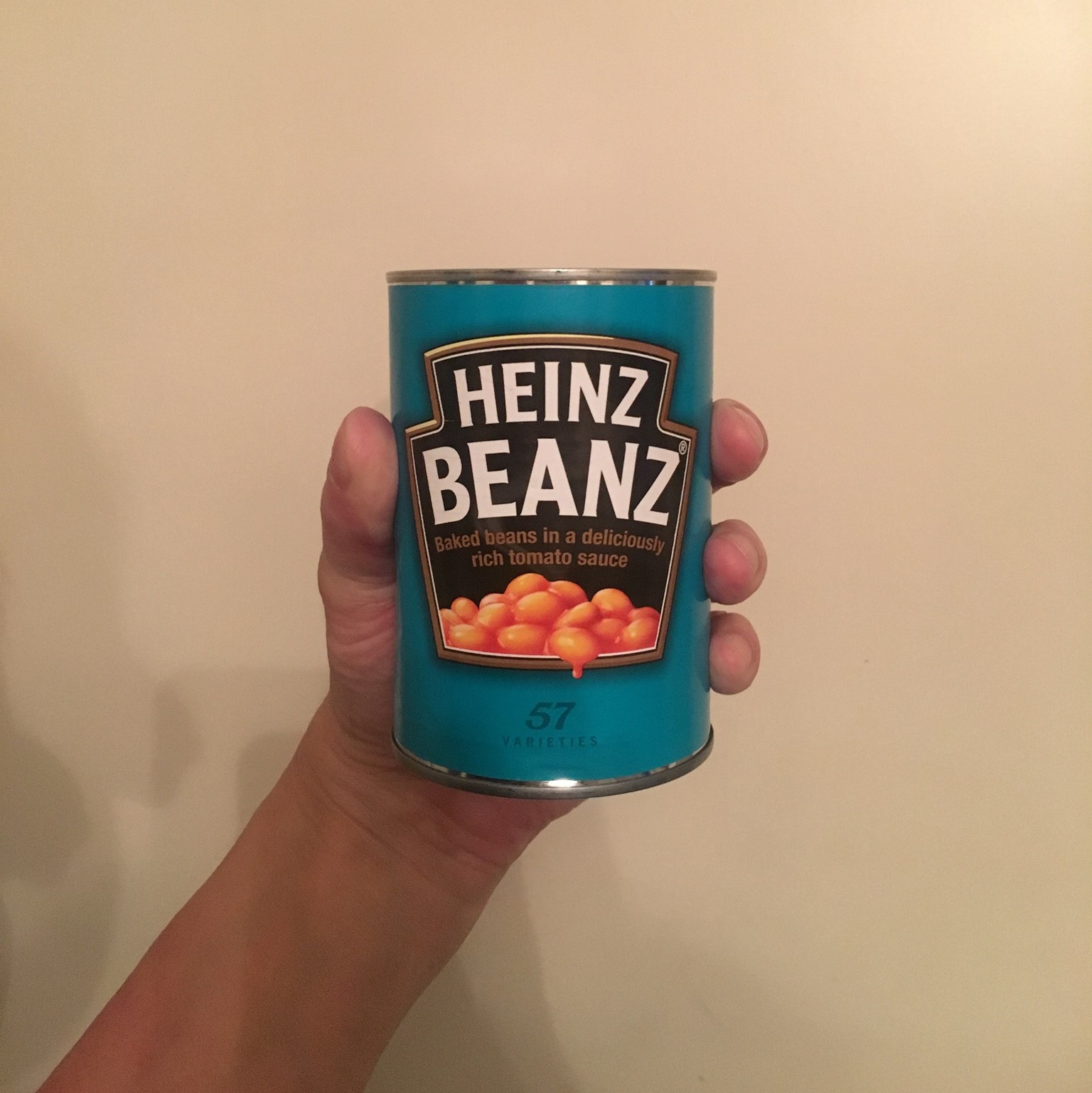}
		\end{subfigure}
		\begin{subfigure}{.495\linewidth}
			\includegraphics[width=\linewidth]{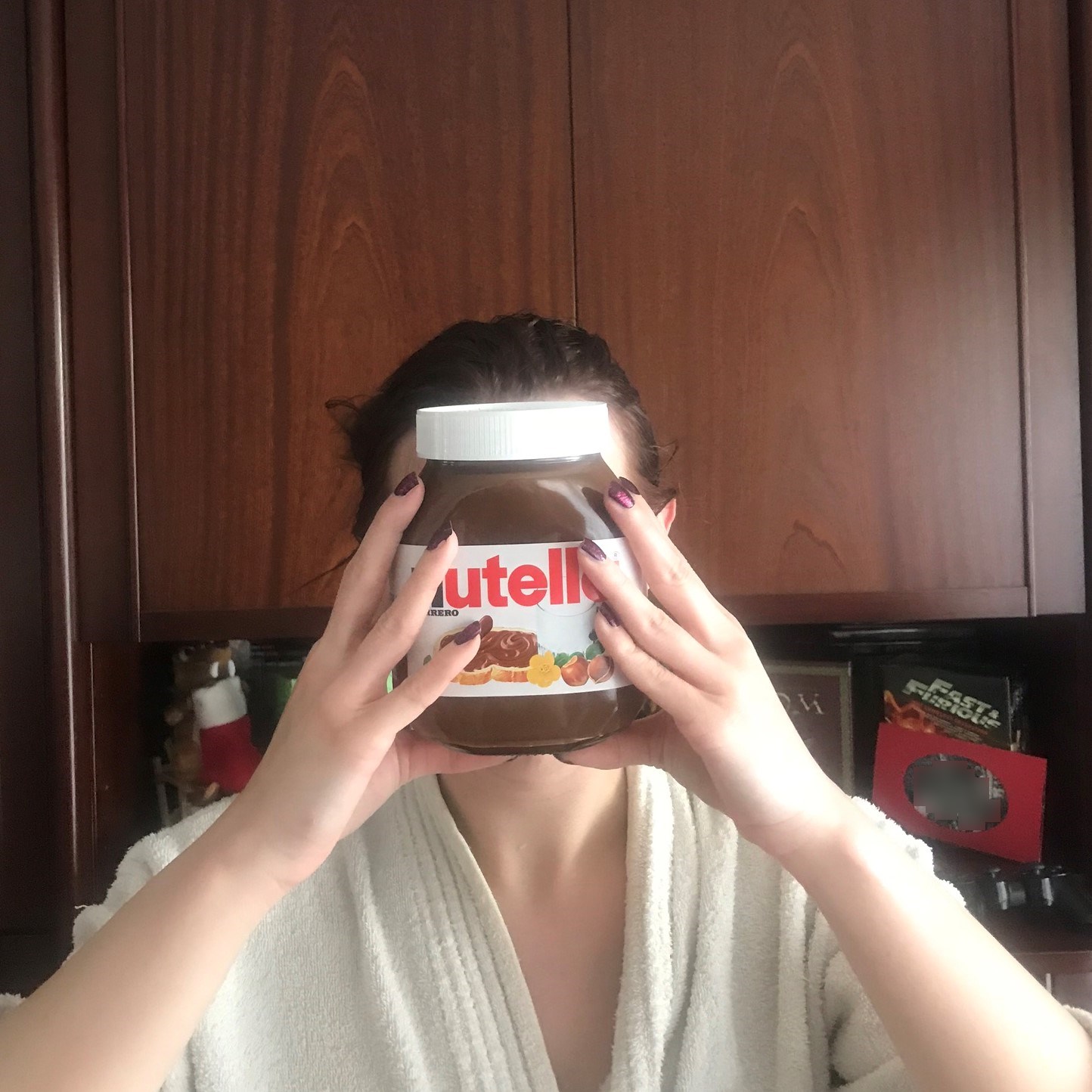}
		\end{subfigure}
	\end{adjustbox}
	\caption{Test images from our dataset. Our model generates \textit{``a female hand holds a can of cocacola above a tiled floor.''}, \textit{``a hand is holding a kinderriegel bar.''}, \textit{``a hand is holding a can of heinz.''}, and \textit{``a young woman is holding a nutella jar in front of her face.''} for the top left, top right, bottom left, and bottom right image, respectively.}
	\label{fig:coke_img_intro}
\end{figure}
Marketing companies have decades of experience in analyzing text fragments in order to investigate sentiment and consumer-brand-relationships.
However, large collections of images in social media rarely come with a description connected to them.
By transcribing images into text, we want to enable marketing companies to rely on text analysis tools that capture two important aspects in descriptions: (1) Occurrence of words that identify a certain brand and (2) attributes and verbs that allow to detect sentiment, affection to a brand and other relevant properties depicted by an image.

Encoder/decoder networks, like the one presented by Vinyals et al.~\cite{vinyals2015show} are promising, when generating captions for input images. Inspired by their work, we build a multi-modal model which simultaneously predicts image ratings. Image ratings describe three different attributes of interactions between humans and branded products in our case. In other words, our model processes three modalities: images, textual descriptions, and image ratings.

In particular, we look at images that contain an object which is related to a brand by depicting a logo of this brand on it. Such an image, for instance, could depict persons drinking out of a Coca-Cola bottle. Figure~\ref{fig:coke_img_intro} shows example images from our test set together with generated captions.

It is of particular interest to us to correctly identify the brand contained in the image, but state of the art models like~\cite{vinyals2015show,vinyals2017show} tend to produce rather generalized descriptions, i.e., the model could just leave out the brand, because it has generalized from pictures of persons holding bottles from different brands. 
For example, the caption generated by \cite{vinyals2015show, vinyals2017show} for the top-left image in Figure~\ref{fig:coke_img_intro} is ``a close up of a person holding a cell phone''. In contrast, we want our model to correctly mention the name of the brand contained in the image within the sentence.

Our second goal is to simultaneously predict attributes that describe the involvement of the human with the brand, whether the branded product appears in a positive or negative context, and whether the interaction is functional or emotional. For example, in a functional interaction a person eats a product, while an emotional interaction would depict people taking selfies with branded products. We call these attributes image ratings, which are integer values from 0 to 4 encoding how much the rating attribute holds. Our goal is to simultaneously predict ratings and, thus, improve the overall quality of the generated sentences in a joint multi-modal optimization.


Our contributions are (1)~We show that using soft targets as a training signal in a classification setting can lead to a better accuracy and model the occasional uncertainty between multiple ground-truth annotators. (2)~We propose a constituent sensitive metric to assess the quality of the subject, predicate, and object of generated sentences. Bundled with synonym tables, we are able to verify the generated captions under the aspect of semantical correctness. We also show that this metric can be calculated on popular image captioning datasets such as MSCOCO~\cite{lin2014microsoft}.

\section{Related Work}

The generation of image descriptions by recurrent neural networks is part of late research. Vinyals et al. \cite{vinyals2015show, vinyals2017show} use this approach in form of an LSTM network. Furthermore, Karpathy et al. use a Bidirectional Recurrent Neural Network \cite{karpathy2015deep} for generating captions. In contrast to Vinyals et al., Karpathy et al. gradually extend their technique to not only describe a whole image, but parts of the image which they call dense captioning~\cite{johnson2016densecap, karpathy2016connecting}.
Kiros et al. \cite{kiros2014unifying} also use an encoder/decoder approach, where they embed images and sentences in the same common space and use an LSTM for encoding the sentences. 


Using soft targets in a classification setting has also been explored by Teney et al.~\cite{teney2017tips}. In a different task called Visual Question Answering, they face a classification problem, where multiple annotators gave an answer to a question regarding the content of an input image. Several thousand unique answers given by the annotators constitute the possible classes. Obviously, annotators do not always agree on the same answer, hence, this classification problem was modeled using soft targets (probability distributions of each answer) given the relative frequency of given answers.

Multi-task learning (MTL) is a long studied domain in machine learning (cf. Caruana~\cite{caruana1993multitask}). Luong et al.~\cite{luong2015multi} work on MTL with focus on sequence to sequence learning. They introduce three different MTL settings for sequence to sequence models, i.e., the one-to-many setting, the many-to-one setting, and the many-to-many setting using multiple decoders and encoders. Our work falls into the category of one-to-many MTL setting, i.e., we use the same encoder (CNN) and multiple decoders (image captioning and three image ratings). 
Harzig et al.~\cite{harzig2018multimodal} use the same configuration, where they generate image captions and image ratings with linear regression at the same time.
\section{Multi-Modal Brand Images Captioning} 
\label{sec:dataset}

Our model uses a real-world dataset, which was created by a market research non-profit association (GfK Verein). The dataset consists of images that depict scenarios of persons interacting with branded products, e.g., a man holding a can with a Coca-Cola logo on it. 
Figure~\ref{fig:coke_img_intro} depicts four sample images from this dataset showing four different classes. For a subset of all images (2,718), five annotators created three image ratings with five possible values ($0-4$) each. 
There are three kinds of image ratings, the first ($r_1$) describing whether the person interacts with the branded product in a positive ($0$) or negative ($4$) way, the second ($r_2$) describing if the person in the image is involved ($0$) with the branded product or uninvolved ($4$), and the third ($r_3$) describing if there is an emotional ($0$) or a functional ($4$) interaction with the branded product. 
Five annotators created a caption for each image, hence, our dataset contains three different modalities with a total of 10,529 images, 52,645 captions and 13,590 image ratings.  
Because of the small number of images in our dataset, we split it into a training and test dataset with a ratio of 9 to 1 and use use 10-fold cross-validation to select the best performing model.


\subsection{Image captioning model}

We build upon the model from Harzig et al.~\cite{harzig2018multimodal}, which use an encoder/decoder network structure similar to Vinyals et al.~\cite{vinyals2015show}. The encoder network is a CNN (\textit{Inception-v3}) that encodes the contents of an image $I$ into a feature map. The decoder network is an LSTM optimized to maximize the probability of generating the ground-truth image caption given an input image. Additionally, they introduce a classification-aware loss function $L_\textrm{cls}$, which penalizes if the correct brand name is not part of the generated caption. Thus, we optimize the loss function
\begin{equation}
\label{eq:lstm_loss}
L(I,S) = - \sum_{t=1}^{N} S_t \log n_t + L_{cls}(I,S),
\end{equation}
where $N$ is the number of words in sentence $S$, $S_t$ is the ground-truth word one-hot vector at time step $t$ and $n_t$ the predicted word probability vector at time step $t$.

\subsection{Soft targets for annotator disagreements}
\label{sec:img_ratings}
For a subset of images, we have three annotations from five different annotators each, rating the interactions between the person and product in the following three dimensions: sentiment (positive vs. negative), involvement (high vs. low), and motive (emotional vs. functional).
The sentiment of the interactions is perceived differently by different annotators.
In order to account for this issue, we propose (1) a majority rating and (2) soft targets for learning to predict the image ratings. Every image rating can have an integer value between 0 and 4. (1) We automatically determined the majority rating, i.e., the rating that most of the five annotators agree on. If there was no majority, we asked an additional annotator to determine this rating based on the image shown. In this case we train with a softmax cross-entropy loss $L_r$ and the majority rating being the ground-truth.
(2) We use soft targets as ground-truth labels. With soft targets, we model the occasional uncertainty between the 5 annotators, e.g., if 4 annotators choose 4 as rating and one annotator chooses 3 as rating, our ground-truth signal $\vec{g}_r$ would be $[0, 0, 0, 0.2, 0.8]$, which describes the probability distribution of the 5 possible values of each rating ($r \in \{r_1,r_2,r_3\}$). We use the sigmoid ($\sigma$) cross-entropy as loss function
\begin{equation}
\label{eq:loss_ratings_sigmoid_ce}
\vec{L}_r(I) = - \vec{g}_r \odot \log(\sigma(\vec{\rho}_r)) - (1 - \vec{g}_r) \odot \log(\vec{g}_r - \sigma(\vec{\rho}_r))
\end{equation}
with $\vec{g}_r$ being the ground-truth and $\vec{\rho}_r$ being the prediction for image rating $r$. Note, that the sigmoid function and the $\log$ are applied element-wise. This can be seen as logistic regression, which predicts the probability of each rating value. The total loss of our model now changes to
\begin{equation}
\label{eq:total_loss_2}
L_{\textrm{total}}=L(I,S) + \vec{1} \cdot \vec{L}_{r_1}(I) + \vec{1} \cdot \vec{L}_{r_2}(I) + \vec{1} \cdot \vec{L}_{r_3}(I).
\end{equation}

\begin{table*}[!htb]
	\centering
	\caption{Results for our models. The first column states the model name, columns 2-4 depict the ratings accuracies for ratings $r_1$, $r_2$, $r_3$. The following 8 columns represent the SPO accuracies which are followed by the BLEU-4 (B-4), METEOR (Met) and CIDEr (Cid) scores. The last two columns show the overall accuraccy (OA) and mean accuracy (MA) as defined in~\cite{harzig2018multimodal}.}
	\label{tab:all-scores}
	\resizebox{1.0\linewidth}{!}{\begin{tabular}{@{}lllllllllllllllll@{}}
			\toprule
			Method (Initialization) & $a_{r_1}$ & $a_{r_2}$ & $a_{r_3}$ & $a_0$ & $a_1$ & $a_2$ & $a_3$ & $a_4$ & $a_5$ & $a_6$ & $a_7$ & B-4 & Met & Cid & OA & MA  \\ \midrule
			MM Regression \cite{harzig2018multimodal} (ft) & - & - & - & - & - & - & - & - & - & - & - & \textbf{0.61} & 0.29 & \textbf{2.11} & 0.91 & 0.83 \\
			\midrule
			base (IC-v3) & 0.57 & 0.51 & 0.55 & 0.00 & 0.76 & 0.83 & 0.68 & 0.80 & 0.66 & 0.68 & 0.59 & 0.53 & 0.33 & 1.56 & 0.75 & 0.56 \\ 
			base-cls (IC-v3 Logos) &0.63 & 0.59 & 0.58 & 0.00 & 0.81 & 0.78 & 0.65 & 0.80 & 0.68 & 0.65 & 0.56 & 0.54 & 0.33 & 1.75 & 0.91 & 0.81  \\ 
			fuse (IC-v3 Logos) & 0.62 & 0.62 & 0.57 & 0.00 & 0.81 & 0.83 & 0.69 & 0.80 & 0.70 & 0.68 & 0.60 & 0.54 & 0.34 & 1.78 & 0.88 & 0.77 \\ 
			fuse-ft (fuse) & \textbf{0.76} & 0.67 & 0.70 & 0.00 & 0.76 & 0.80 & 0.63 & 0.81 & 0.67 & 0.66 & 0.56 & 0.51 & 0.33 & 1.62 & 0.77 & 0.46 \\ 
			\midrule
			soft targets (base-cls) & 0.63 & 0.59 & 0.58 & 0.00 & 0.81 & 0.78 & 0.65 & 0.80 & 0.68 & 0.65 & 0.56 & 0.54 & 0.33 & 1.75 & 0.91 & 0.81 \\ 
			\midrule
			soft targets (fuse-ft) & \textbf{0.76} & 0.70 & \textbf{0.71} & 0.00 & 0.85 & 0.87 & 0.76 & \textbf{0.86} & 0.75 & 0.76 & 0.68 & \textbf{0.61} & 0.37 & 2.04 & 0.90 & 0.79 \\ 
			soft targets-ft (fuse-ft) & 0.75 & \textbf{0.71} & \textbf{0.71} & 0.00 & \textbf{0.86} & \textbf{0.89} & \textbf{0.78} & \textbf{0.86} & \textbf{0.76} & \textbf{0.78} & \textbf{0.70} & \textbf{0.61} & \textbf{0.38} & 2.07 & 0.91 & 0.80 \\ 
			ratings cls (fuse-ft) & 0.73 & 0.68 & 0.68 & 0.00 & 0.84 & 0.88 & 0.75 & 0.84 & 0.72 & 0.75 & 0.65 & 0.60 & 0.37 & 2.04 & 0.91 & 0.82 \\ 
			ratings cls-ft (fuse-ft) & 0.75 & 0.70 & 0.67 & 0.00 & \textbf{0.86} & 0.88 & 0.77 & 0.85 & 0.75 & 0.76 & 0.67 & \textbf{0.61} & 0.37 & 2.06 & \textbf{0.92} & \textbf{0.84} \\ 
			\midrule
			gt & - & - & - & - & 0.68 & 0.98 & 0.67 & 0.98 & 0.67 & 0.97 & 0.66 & 0.52 & 0.38 & 1.94 & - & - \\
			\bottomrule
		\end{tabular}
	}
\end{table*}
\section{SPO Captioning Metrics}

Common metrics developed for the task of machine translation are BLEU~\cite{papineni2002bleu} and METEOR~\cite{banerjee2005meteor}. 
CIDEr~\cite{vedantam2015cider} is a metric developed specifically for image captioning and designed to correlate well with human judgment~\cite{vedantam2015cider}.
All these metrics have shown to score higher for machine-generated sentences than for human-generated sentences for the MSCOCO captioning challenge in some cases. We also find that captions generated by our models score higher in comparison to the ground-truth sentences (see Section~\ref{sec:res_common_metrics} and Table~\ref{tab:all-scores}). Common machine translation metrics also have the downside that they do not capture tiny important pieces of generated sentences like the object of interest or the predicate. For example, the generated sentence \textit{A \textbf{male} hand holds a can of cocacola above a tiled floor.} for the ground-truth sentence \textit{A \textbf{female} hand holds a can of cocacola above a tiled floor.} has a BLEU-4 score of 0.827, which is very high. In our setting such minor differences are very important and, thus, we introduce novel metrics and make the assertion that popular metrics may disregard the semantics of the captions.

To allow for a more fine-grained evaluation of generated captions than existing methods do, we introduce subject-predicate-object (SPO) accuracies. We show in Section~\ref{sec:res_spo_on_mscoco} that we can also use this metric on the MSCOCO~\cite{lin2014microsoft} image captioning dataset, therefore, it is presenting itself as an alternative to the common metrics that try to measure the quality of generated sentences.
For our dataset, we manually collect the subject, predicate, and object of each of our ground-truth sentences. 
Since the brand names of the objects on the images are already known, we made sure that the annotators do not choose the brand name as object, but the actual object, i.e., the caption ``A hand is holding a Coca Cola can in a car.'' results in the SPO triple \textit{(hand, hold, can)}.

Because we have five sentences from different annotators per image, we also get 5 ground-truth SPO triples per image. We require the LSTM generated sentence to only match one of the SPO triples. We define three different matching criteria $m_n$ per generated sentence: 
(1) we set $m_s=1$, if the subject of the generated sentence matches, (2) $m_p = 1$, if the predicate matches and (3) $m_o=1$ if the object matches. $m_s, m_p, m_o$ are set to $0$, otherwise. By combining those matching criteria, we define eight derived matchings $m_0:=\neg m_p \wedge \neg m_o \wedge \neg m_s$, $m_1:=m_o$, $m_2:=m_s$, $m_3:=m_o \wedge m_s$, $m_4:=m_p$, $m_5:=m_p \wedge m_o$, $m_6=m_p \wedge m_s$ and $m_7:=m_p \wedge m_o \wedge m_s$.
For example, $m_4$ describes whether the predicate was generated correctly and $m_7$ equals $1$ if subject, predicate, and object were generated correctly. 
We define the accuracies $a_n$ based on those matching criteria to be the fraction of generated captions over the test set, which satisfy the matching criterion. Thus, we have a number of different accuracies, which tell us how often we generated a sentence with the correct subject, predicate, or object, and all combinations of these.
Different matching accuracies have different evaluation emphases. For example, for a captioning task that focuses on the interactions between persons and objects the $a_4$ accuracy may be of special interest, while for a task that specializes on correctly identifying the actor, the $a_2$ accuracy may be suited best.

Since two different words can literally have the same meaning (e.g., \textit{ad} and \textit{advertisement}), we use synonym tables for our subjects and objects. Based on an analysis of captions collected by our annotators, we found that annotators tend to avoid repeating words (e.g., they alternate between pack, package and packaging). Hence, we created synonym tables consisting of manually encoded bidirectional and unidirectional synonyms. Bidirectional synonyms are of equal meaning. 
Unidirectional synonyms can not be used synonymous in all contexts, e.g., the words \textit{man} and \textit{boy} can be replaced by \textit{guy}, but we can not implicitly infer the age of a \textit{guy}.

\section{Results}
\label{sec:results}

We conducted a series of experiments on different models. In this section, we present results for 9 models, which we depict in Table~\ref{tab:all-scores}. All models are trained using the classification-aware loss (see~\cite{harzig2018multimodal}). Model \textit{base} is a model initialized with parameters from a vanilla \textit{Inception-v3} network. \textit{base-cls} is initialized with an \textit{Inception-v3} network finetuned to classify our 26 logo classes. \textit{fuse} is initialized the same way, but trained on the LogosExtended dataset (logos + MSCOCO) with the CNN network freezed, while \textit{fuse-ft} is trained for 2M additional iterations with the \textit{Inception-v3} CNN unfreezed. Our final models \textit{soft targets}, \textit{ratings cls} and the corresponding finetuned models (ft) use \textit{fuse-ft} as initialization and the logos only dataset. To show the effectiveness of MTL, we also trained a model with \textit{base-cls} as initialization (\textit{soft targets (base-cls)}) without unfreezing the encoder network. All scores except the sentence classification accuracy decrease considerably when compared to a model trained in a multi-task setting (e.g., \textit{soft targets (fuse-ft)}).
\subsection{Training}
%
We train the captioning model with the same parameters Vinyals et al.~\cite{vinyals2015show,vinyals2017show} use. We train our logos only dataset for 68 epochs and decay the learning rate 8 times by a factor of $0.5$. We train the \textit{fuse} model with the LogosExtended dataset, where we multiply our dataset 8 times to compensate for the fewer examples compared to the MSCOCO dataset. 
We train \textit{fuse} with the extended dataset for 56 epochs and decay the learning rate 7 times. For our extended model with the classification-aware loss, we just optimize all losses simultaneously according to Equation~\ref{eq:total_loss_2}. In this step, we freeze the \textit{Inception-v3} parameters.

%
\subsection{Image ratings}
\label{sec:res_img_ratings}

\begin{figure*}[!htb]
	\centering
	\includegraphics[width=\linewidth]{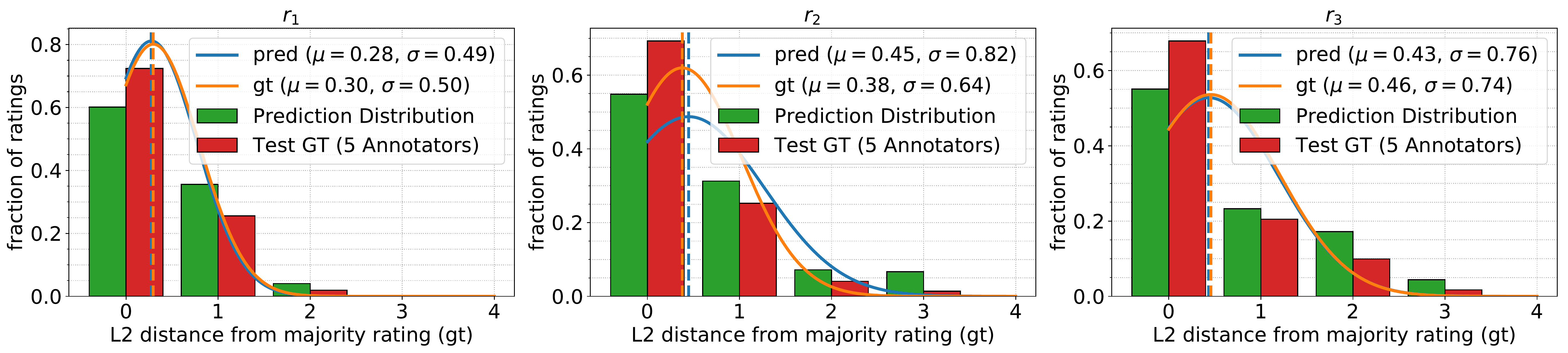}
	\caption{The predicted (green) L2 deviations from the ground-truth majority rating in comparison to the mean L2 deviations (red) of the annotator labels to the majority rating.}
	\label{fig:ratings_dists}	
\end{figure*}
In the following, we present the results for predicting our three different image ratings $r_1$, $r_2$ and $r_3$. For each rating, we use FC layers with 5 output neurons and model it (1) with a softmax cross-entropy loss as a classification problem and (2) with a sigmoid cross-entropy loss as a logistic regression problem, i.e., we predict the probality distribution among the different rating values (an integer-value between $0$ and $4$).
We evaluate the performance by using the accuracy measures $a_{r_1}$, $a_{r_2}$ and $a_{r_3}$, which tell us how often our model predicts the correct rating out of all evaluation examples. 
To allow comparison between (1) and (2), we use the majority rating as ground-truth for both methods. For the predicted value, we take the argmax value of the probability distribution vector in case of (2) and the argmax of the softmax for method (1) as is common practice. As we see in Table~\ref{tab:all-scores}, our best model \textit{soft targets-ft} achieves 0.75, 0.71 and 0.71 for the accuracies $a_{r_1}$, $a_{r_2}$ and $a_{r_3}$, respectively. For the top left image in Figure~\ref{fig:coke_img_intro} the image ratings predict that the interaction is neither positive nor negative ($r_1=2$), the person is rather involved with the product ($r_2=1$) and the interaction is more functional than emotional ($r_3=3$).

In addition to analyzing the image ratings with the accuracy measure, we also compare the predicted probability distribution (2) against the distribution generated by the annotators, i.e., we want to examine if we can imitate the occasional uncertainty between our human annotators. We calculate the L2 distance from the predicted value to the majority rating for both the ground-truth values and predicted values. 
In Figure~\ref{fig:ratings_dists}, we visualize the prediction distribution of model \textit{soft targets-ft} compared to the test split ground-truth. The green bars show the fraction of predicted ratings, which have an L2 distance to the ground-truth rating of $0 - 4$, while the red bars show the L2 distance of each annotator to the majority rating. We can observe that the predicted distribution closely models the annotator disagreement on the hold-out test set. We also visualized the normal distributions of the predictions and ground-truth and their mean and standard deviations. For $r_1$ the annotators deviate $\mu = 0.30$ from the majority rating on average and have a standard deviation of $\sigma=0.50$. The distribution predicted by our model comes really close with $\mu = 0.28$ and $\sigma = 0.49$. Also, rating $r_3$ has a very similar distribution and only $r_2$ differs slightly from the ground-truth.

Harzig et al.~\cite{harzig2018multimodal} face a similar problem, and they model image ratings as linear regression which directly predicts the rating value as a float value. They evaluate the predictions by calculating the L2 distance from the float value prediction to the mean of the ground-truth annotations. To compare the predicted deviations, they also calculate the deviation from one annotator to the mean of the others and average this value for all annotators. The best model in their work (see the first row of Table~\ref{tab:all-scores}) achieves mean L2 distances of 0.52, 0.78, and 0.70 for ratings $r_1$, $r_2$, and $r_3$, respectively.  The mean L2 distances for the ground-truths (they compare one annotator against the mean of the other 4 and average over all 5 annotators) are $0.25$, $0.19$, and $0.31$, i.e., the mean of the predictions differ from the ground-truth by $0.27$, $0.59$, and $0.39$, respectively. In comparison, our best model only differs from the ground-truth mean by $0.02$, $0.07$, and $0.03$ for ratings $r_1$, $r_2$, and $r_3$, respectively.

For a thorough analysis, we employed the classical classification approach, i.e., we used a softmax cross-entropy loss for the image ratings with the majority rating as the ground-truth.
In Table~\ref{tab:all-scores}, we denote the majority rating sampling strategy as \textit{ratings cls} and \textit{ratings cls-ft}. None of those experiments could surpass the classification accuracy of the soft targets approach ($0.75$, $0.70$, and $0.67$ vs. $0.75$, $0.71$, and $0.71$ for rating accuracies $a_{r_1}$, $a_{r_2}$, and $a_{r_3}$, respectively).

\subsection{SPO accuracies}
In Table~\ref{tab:all-scores}, we also report the scores our models achieve with our proposed SPO accuracy metrics. As we did for all other evaluation measures, we calculated the ground-truth accuracies for $a_0$ - $a_7$ by calculating the accuracies for every annotator against the other four annotators and averaged over the accuracies. In our models we use beam search with a beam size of $k=3$ during evaluation and inference. During evaluation, we choose the caption which matches most of the three sentence clauses defined by us (subject, predicate, object). With the \textit{soft targets-ft} model, we achieve the best scores, i.e., we identify the correct predicate in 86\% of all cases and generate completely correct sentences (according to the SPO metric) in 70\% of all cases. In comparison, our base model only generates sentences with matching subjects, predicates and objects only in 59\% of all cases (\textit{base}). Note, that $a_0=0$ represents a perfect accuracy for the accuracy $a_0$. This means that for every generated caption at least one matching clause was generated.

\subsection{Sentence classification accuracy}
We employ the SCA metrics from~\cite{harzig2018multimodal}, which we use as an indicator on how well our model performs in terms of generating the correct brand name. Compared to \cite{harzig2018multimodal}, we achieve similar performance with our model \textit{soft targets-ft} and better performance when using the traditional classification approach for the image ratings (\textit{ratings cls-ft}). Unsurprisingly, the finetuned models (\textit{\dots-ft}) perform better than the base model.
\subsection{Common captioning metrics}
\label{sec:res_common_metrics}
Additionally, we also evaluate our models with common machine translation and image captioning metrics. We use the official script provided by MSCOCO to calculate these metrics. We report ground-truth scores in the last row of Table~\ref{tab:all-scores}. We calculated them by comparing each annotator against the other four and then averaging over all of them.
We only report scores calculated on our dataset without MSCOCO.
Hence,  when training on both datasets at once (\textit{fuse-ft}), the scores drop in comparison to training on the logos dataset only (see method \textit{base-cls}). 
For final optimization we use the \textit{fuse-ft} as initialization for the logos only dataset. We see that this model (\textit{soft targets-ft}) performs best for the common sentence metrics except for the CIDEr metric. Thus, we conclude that using soft targets instead of linear regression (cf. Harzig et al.~\cite{harzig2018multimodal}, \textit{MM Regression}) has no negative effect on the overall sentence quality. Using the traditional classification approach instead of soft targets (\textit{ratings cls-ft}) yields similar results.

\subsection{SPO accuracies on MSCOCO}
\label{sec:res_spo_on_mscoco}
\begin{table}[!htb]
	\centering
	\caption{SPO accuracies for the Show and Tell model on the MSCOCO development test split.}
	\label{tab:spo_mscoco}
	\resizebox{\linewidth}{!}{\begin{tabular}{@{}lllllllll@{}}
			\toprule
			& $a_0$ & $a_1$ & $a_2$ & $a_3$ & $a_4$ & $a_5$ & $a_6$ & $a_7$ \\ \midrule
			NICv2 \cite{vinyals2017show} & 0.00  & 0.63  & 0.67  & 0.42  & 0.64  & 0.37  & 0.39  & 0.22  \\ \bottomrule
	\end{tabular}}
\end{table}
We want show that our SPO accuracies measure can be used on other image captioning datasets. To do so, we generated SPO ground-truth triples from the human-annotated sentences of the MSCOCO validation split. We use the natural language processing library spaCy\footnote{\url{https://spacy.io/}} to extract SPO triples from sentences similar to how our human annotators extracted the triples from our dataset and publish these annotations\footnote{\url{https://github.com/philm5/mscoco-spo-triples}}. We then trained the Show and Tell~\cite{vinyals2015show,vinyals2017show} implementation from the TensorFlow models repository on MSCOCO, which yielded scores slightly worse than those published by Vinyals et al.~\cite{vinyals2017show}. \\
We found that the ground-truth captions of MSCOCO often contain no predicate (e.g., \textit{A room with blue walls and a white sink and door.}). We were able to automatically extract SPO ground-truth triples (including the triples without a predicate) from 184,308 out of 202,654 image captions in total. Additionally, we did not collect synonym tables for the MSCOCO ground-truth SPO triples, which decreases the final SPO accuracies. In Table~\ref{tab:spo_mscoco}, we list the different accuracies the MSCOCO model achieves with our metric. We match at least one object in 63.14\%, the subject in 66.82\%, and the predicate in 63.91\% of all cases. However, as we see with the $a_7$ accuracy, the Show and Tell model only produces captions which contain the correct subject, predicate and object only in 21.97\% of all cases. Manually annotated SPO triples would lead to a more accurate result.


\section{Conclusion}

We presented an architecture capable of simultaneously generating image captions and ratings for images depicting scenes of interactions between humans and branded products. Our focus lies on generating captions that satisfy other constraints than traditional image captioning pipelines. First, we presented novel clause-focused accuracy metrics that focus on the correct transcription of the subject, object, and predicate. In addition to the ground-truth image captions, we annotated subject, object, and predicate (including synonyms) for our dataset to be able to measure the quality of our generated sentences in terms of these important clauses of the sentence. Furthermore, we automatically extracted subject, predicate, and object from the ground-truth annotations of the popular MSCOCO dataset to verify our approach by applying our metrics on a different dataset. Second, in a multi-modal training procedure, we predicted ratings from the input image, which describe (1) if the person is involved with the brand, (2) whether the interaction is rather positive or negative, and (3) whether the interaction is functional or emotional. By modeling the occasional uncertainty between annotators with a soft target logistic regression, we were able to improve overall sentence quality in an end-to-end multi-task optimization.

\section*{Acknowledgment}

This work was funded by GfK Verein. The authors would like to thank Holger Dietrich, Raimund Wildner and Andreas Neus for the great collaboration.



%
%
%
\bibliographystyle{IEEEtran}
\bibliography{IEEEabrv,mipr2019}

\end{document}